\documentclass[a4paper]{llncs}	%% wird das Dokument in Deutsch verfasst, so muss der Parameter ``deutsch'' in den eckigen Klammern eingetragen werden: \documentclass[deutsch]{llncs}

\usepackage{graphicx}
\usepackage{todonotes}
\usepackage{setspace}

\begin{document}

{
\begin{figure}
  \includegraphics[width=0.35\textwidth]{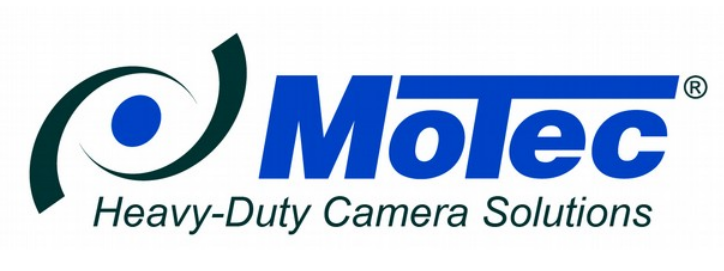}
  \hfill
  \includegraphics[width=0.35\textwidth]{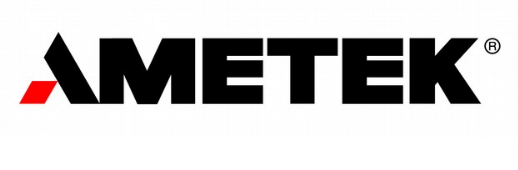}
\end{figure}

\begin{center}
\begin{doublespace}
\textbf{\begin{Large}Mixing Real and Synthetic Data to Enhance Neural Network Training - A Review of Current Approaches\end{Large}}
\end{doublespace}
\end{center}
\vfill
\begin{figure}
  \includegraphics[width=0.3\textwidth]{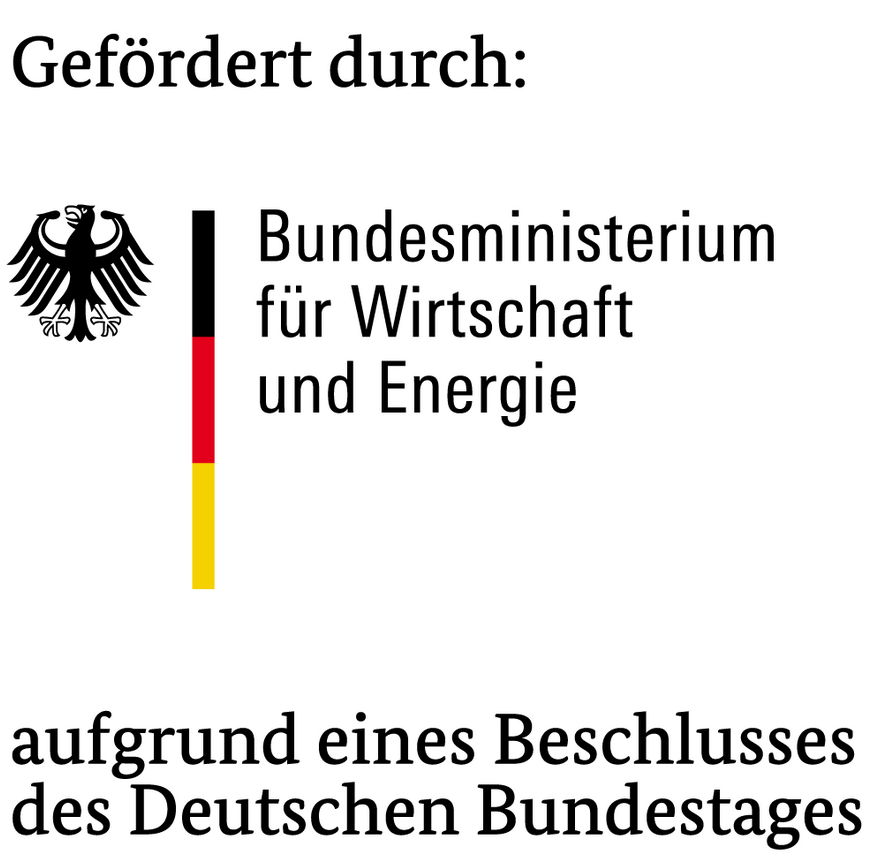}
  %\hfill
  %\includegraphics[width=0.6\textwidth]{images/gama_logo.png}
\end{figure}

\noindent Das diesem Bericht zugrundeliegende Vorhaben wurde mit Mitteln des Bundesmi\-nisteriums f\"ur Wirtschaft und Energie unter dem F\"orderkennzeichen 19A16015D gef\"ordert. Die Verantwortung f\"ur den Inhalt dieser Ver\"offentlichung liegt beim Autor/bei der Autorin.

\vspace{0.5cm}
\noindent This work has been funded by the German Ministry for Economic Affairs and Energy (reference number: 19A16015D).
\vspace{0.5cm}
}

\title{Mixing Real and Synthetic Data to Enhance Neural Network Training - A Review of Current Approaches}

 \author{Viktor Seib, Benjamin Lange, Stefan Wirtz}
 \institute{Motec GmbH, Oberweyerer Stra\ss e 21, 65589 Hadamar-Steinbach, Germany \\
 \email{\{viktor.seib, benjamin.lange, stefan.wirtz\}@ametek.com}}

%\author{names hidden for blind review}
%\institute{institue hidden for blind review \\
%\institute{Technische Universität Kaiserslautern, Gottlieb-Daimler-Straße, 67663 Kaiserslautern
%\and Springer Heidelberg, Tiergartenstr. 17, 69121 Heidelberg, Germany\\
%\email{\{mail hidden for blind review\}}}

\maketitle

\begin{abstract}
Deep neural networks have gained tremendous importance in many computer vision tasks.
However, their power comes at the cost of large amounts of annotated data required for supervised training.
In this work we review and compare different techniques available in the literature to improve training results without acquiring additional annotated real-world data.
This goal is mostly achieved by applying annotation-preserving transformations to existing data or by synthetically creating more data.
\end{abstract}

\begin{keywords}
Convolutional Neural Networks, Transfer Learning, Data Augmentation, Synthetic Data, Review
\end{keywords}

\section{Introduction}

Many tasks in object classification and detection can nowadays be performed by computers with a human level accuracy \cite{russakovsky2015imagenet}.
These advances in computer vision are achieved by creating large annotated datasets for the targeted application domain.
However, not all domains allow the collection of large amounts of data.
This is the case if some rare events need to be detected with high confidence.
Another example are privacy issues that impede data collection for health care applications.
Indeed, recent research indicates that the current limitations to the performance of convolutional neural networks is not the network architecture, but the limited amount of available data for training \cite{drivingmatrix}.
Due to the enormous work needed to annotate data, nowadays large-scale datasets exist only for a limited range of applications \cite{deng2009imagenet}, \cite{lai2011large}, \cite{shahroudy2016ntu}.
Consequently, researchers are interested in methods to create powerful networks with less training data -- or at least with less real-world data.

The motivation to use less real-world data is, on one hand, the expensive acquisition of the data and, on the other hand, the even more expensive annotation.
In contrast, synthetic data can be generated in arbitrary amounts once a suitable framework is properly setup.
Even better: the annotations are generated along with the data itself at (almost) no additional cost.
Depending on the application domain and size of the artificial dataset it can be used as a large dataset for transfer learning or even to train a neural network from scratch.
There are also some downside to consider.
Neural networks learn a latent feature distribution of the presented training data.
Thus, the resulting network will exhibit a poor performance if the synthetic data does not properly reflect the feature distribution of the data in the target domain.
This performance gap is referred to as domain shift and is a common problem when training with synthetic data \cite{sankaranarayanan2018learning}.

In this work we focus on the domain of urban and traffic scenes.
The goal is to review convolutional neural network training scenarios that improve the network's performance without additional real-world data.
Please note that this review is not exhaustive.
There is a lot of research currently done in this field and not all approaches, or even groups of approaches, are equally well treated in this review.
Nevertheless, we hope that the exemplary approaches compiled in this review will provide further insights and open new directions for investigations.

We start off by reviewing successful methods for data augmentation and some common approaches in transfer learning in Section\,\ref{littledata}.
This includes data augmentation methods that were used in prevalent and successful neural networks.
We also discuss different options typically used in transfer learning and fine-tuning and provide a short note on the feedforward design to calculate network weights.
Section\,\ref{syndata} is the main part of this work.
Therein we present synthetic urban scene datasets that are currently applied in research.
We discuss the training procedures and experiments performed by the authors of the respective datasets.
Additionally, a short section is dedicated to some examples to diminish the effects of domain shift.
There is one type of synthetic datasets that nowadays is still not available to the research community: synthetic GAN-generated, photo-realistic images of urban scenes.
In Section\,\ref{outlook} we review some work into that direction and briefly state why such datasets might be expected in near future.
Finally, Section\,\ref{summary} concludes this paper and summarizes some messages learnt from the reviewed approaches.

%%%%%%%%%%%%%%%%%%%%%%%

\section{Little Data - Smartly Used}\label{littledata}

This section introduces common approaches to reuse trained networks available online for other purposes.
Further, we review methods to artificially increase the amount of data available for training without collecting or annotating additional images.

\subsection{Pretrained networks}

Pretrained network weights for all commonly known network architectures and deep learning frameworks are available online for free.
These networks are mostly pretrained on large datasets such as ImageNet \cite{deng2009imagenet}, Pascal VOC \cite{everingham2015pascal} or COCO \cite{lin2014microsoft}.
The size of these datasets allows the neural networks to learn abstract representations for many different object classes and embed generalized feature representations.

When adapting a neural network to a custom use case, the size of the use case-specific dataset often will be smaller than the size of the dataset used for pretraining.
Therefore, for a customized application based on neural networks one should first check which available network architecture best meets the requirements regarding accuracy and available computational resources.

In most cases it is advisable to take an existing, pretrained network as the basis for a new custom application.
Surprisingly, this is even the case when the data type and application domain of the target network significantly differ from the original pretrained network.
For instance Schwarz et al. \cite{schwarz2015rgb} and Eitel et al. \cite{eitel2015multimodal} use a network pretrained on ImageNet with RGB images to train a network on range image data.
Another example presented in \cite{esteva2017dermatologist} shows the successful application of a network pretrained on ImageNet for skin cancer classification.

Starting with a pretrained network is the simplest method to reduce the amount of necessary data for training.
Once a pretrained network is chosen, commonly transfer learning and fine-tuning are applied to adapt the network to a specific target domain.

\subsection{Transfer Learning and Fine-Tuning}

While the terms are often used interchangeably, transfer learning and fine-tuning are different techniques that are often applied simultaneously.
Both methods are used to transfer the trained weights from one application domain to another \cite{yosinski2014transferable}, \cite{oquab2014learning}.
For example, a network trained for autonomous driving in urban scenes can be applied for autonomous navigation in a harbor environment.

The last layer of a network is responsible for the trained task, for example classification or detection.
The last layer also determines the total number of classes the network can distinguish.

In transfer learning we replace the last layer by a layer with a desired number of classes.
This modified network is then trained with the application-specific dataset.
All other layers of the network are ``frozen'' - this means their weights are kept constant and are not changed by the training procedure with the application-specific dataset.

The term fine-tuning means that the weights in all or some of the other layers are not frozen.
The subsequent training with the custom dataset therefore fine-tunes the existing weights to the new domain.
Since the more shallow layers learn low-level features that differ less across domains, a smaller learning rate should be used than for deeper layers of the network.
If shallow layers are changed too quickly or too much, the network might overfit the training data of the target dataset and lose its generalization ability.

Some techniques used to train neural networks from scratch can be readapted to transfer learning.
In \cite{simonyan2014very} Simonyan et al. describe the training of VGG - one of the commonly used network architectures.
Initially, a shallow version of the network was created and trained on the data.
The network was then extended and the previously trained weights used for initialisation of the shallow layers.

A similar strategy can be applied for transfer learning.
A shallow network corresponding to the last few layers of the target pretrained network can be trained from scratch on the application specific dataset.
Then, instead of only replacing the last layer, the last layers are replaced by the shallow network with its weights.
Finally, the whole network is further trained and fine-tuned on the target dataset.

\subsection{Data Augmentation} \label{data-augment}

Data augmentation is a common method to artificially increase the size of an existing dataset \cite{mikolajczyk2018data} and is often combined with transfer learning and fine-tuning.
While there are different types of methods \cite{shorten2019survey}, they all generate altered data from the existing set of training images.

\begin{figure}
 \centering
 \includegraphics[width=0.7\textwidth]{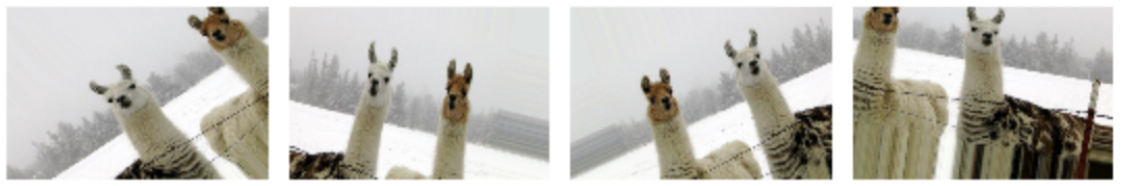}
 \caption{Basic image manipulations for data augmentation. Image taken from \cite{perez2017effectiveness}.}
 \label{fig:basic_manipulations}
\end{figure}

Most commonly, basic image manipulations such as horizontal flipping, random cropping or rotating are used (Figure\,\ref{fig:basic_manipulations}).
Other methods such as normalization (mean subtraction), color space transformations, photometric distortions and noise injection are also widely used.
In general, any alteration of the input image that does not change the label information is acceptable.
For instance, horizontally flipping an image of a cat still produces a cat.
While applying the same transformation to an image of the digit five produces an image with an invalid number that must not be trained as a five.

In the training of all the famous architectures, such as AlexNet \cite{krizhevsky2012imagenet}, InceptionV3 \cite{szegedy2015going}, VGG \cite{simonyan2014very} and ResNet \cite{he2016deep}, mean subtraction and random cropping were used to augment the training set.
While for AlexNet random cropping was used with a fixed size, in the training of all other architectures images were cropped with a random size and different aspect ratios.
Out of these networks, Inception was the only one without horizontal flipping augmentation.
Contrary, the augmentations for Inception included photometric distortions and random changes in brightness, contrast, color and even different interpolation methods while cropping images with different sizes.
Out of these transformations, only random brightness was applied to AlexNet, while random changes in color were used for VGG and ResNet.

Data augmentation is usually applied at training time to enable the network to generalize better by presenting it different variations of the data.
However, when the networks mentioned above entered the ImageNet large scale visual recognition challenge \cite{russakovsky2015imagenet} they also applied augmentation at test time.
Test time augmentation does not affect the weights of the network.
The purpose is to present the network with different crops and transformations of the query image and average the results.
This is a viable way for offline computations for a competition.
In practical applications with possible real-time constraints this might not be feasible.

Random erasing \cite{zhong2017random} is another strategy for augmentation.
A small part (random size and position) of the input image is replaced by a solid color or noise to simulate occlusion.
This technique has to be used carefully in order not to completely erase the pixels of the image that constitute the object described by its label.

Other methods for data augmentation are less intuitive to humans.
For example, in sample pairing \cite{inoue2018data} two random images of different classes are averaged along each of the RGB channels.
The resulting image gets the label of the first randomly selected image.
Surprisingly, this data augmentation strategy was reported to reduce the error rate \cite{inoue2018data}.

Generative adversarial networks (GANs) \cite{goodfellow2014generative} have also been applied for data augmentation.
GANs have gained recent attention in many fields, for example the generation of photo-realistic face images \cite{karras2017progressive} or style transfer \cite{johnson2016perceptual}.
In the same manner as a GAN can be trained to generate faces, it can also be trained to generate images to extend a dataset.
This was demonstrated in the domain of medical images by Maayan et al. in \cite{frid2018gan}.
However, training a GAN requires a large amount of images.
If the available dataset is small it will not suffice to train a GAN and traditional data augmentation techniques should be used.

\begin{figure}
 \centering
 \includegraphics[width=0.7\textwidth]{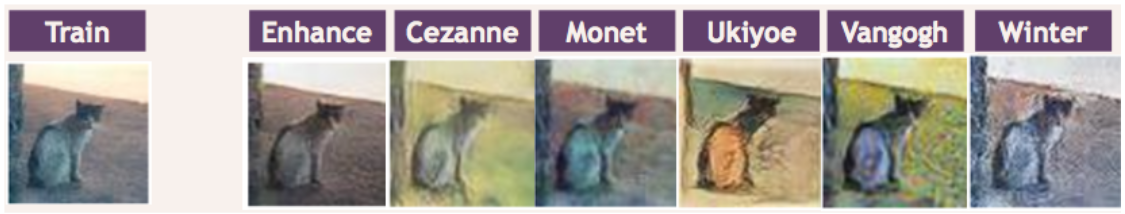}
 \caption{Illustration of style transfer applied to an image. Image taken from \cite{perez2017effectiveness}.}
 \label{fig:style_images}
\end{figure}

On the other hand, many implementations of style transfer GANs are available online, including pretrained weights.
These can be applied for data augmentation purposes as suggested for example in \cite{perez2017effectiveness}.
The resulting images show the same content, but in a different artistic style (Figure\,\ref{fig:style_images}).
This type of data augmentation is similar as generating synthetic data using domain randomization (see Section\,\ref{examp-domain}).

Another important augmentation method using GANs is the possibility to transfer images from one domain to another \cite{zhu2017unpaired}.
Of special interest for autonomous driving would be transferring images between day and night, summer and winter and sunny and rainy weather \cite{shorten2019survey}.
This method does not need paired image examples, but still requires a sufficiently large dataset to train domains of interest.
We hope to see pretrained weights available online for corresponding domain translation models in the near future.

\subsection{Feedforward Design}

A recent paper proposes a completely different approach to compute network weights that deviates from traditional training schemes.
The so called \textit{feedforward design} omits the backpropagation algorithm used to train neural networks in favor of an algebraic approach that only requires one forward pass \cite{kuo2019interpretable}.
The benefits reported are faster training, less data required for training and less vulnerability to adversarial attacks.
The results presented in \cite{kuo2019interpretable} were applied only on simple datasets and more research is necessary to fully assess the advantages one could gain using that approach.
We imagine that future work could result in some efficient pipelines to generate initial weights for new network architectures that lack pretrained weights.

%%%%%%%%%%%%%%%%%%%%%%%%%%%%%%%%%%%%%%%%%%%%%%%%%%%%%%%%%%%%%%%%%%%%%%%%%%%%%%%%%%%%%%%%%%%%%%%%%%%%%%%%%%%%%%% 
%%%%%%%%%%%%%%%%%%%%%%%%%%%%%%%%%%%%%%%%%%%%%%%%%%%%%%%%%%%%%%%%%%%%%%%%%%%%%%%%%%%%%%%%%%%%%%%%%%%%%%%%%%%%%%% 
%%%%%%%%%%%%%%%%%%%%%%%%%%%%%%%%%%%%%%%%%%%%%%%%%%%%%%%%%%%%%%%%%%%%%%%%%%%%%%%%%%%%%%%%%%%%%%%%%%%%%%%%%%%%%%% 
%%%%%%%%%%%%%%%%%%%%%%%%%%%%%%%%%%%%%%%%%%%%%%%%%%%%%%%%%%%%%%%%%%%%%%%%%%%%%%%%%%%%%%%%%%%%%%%%%%%%%%%%%%%%%%% 

\section{Synthetic Data}\label{syndata}

% allgemeine motivation

The previously discussed methods aim at altering and augmenting the available real-world data to artificially increase the dataset size.
Naturally, this real-world data is hard to collect and time consuming to annotate.

In contrast to previous methods, the methods discussed here create synthetic data for the specific application domain.
Scenarios specific to an application domain can be modelled with 3D engines used to design computer games such as the Unreal Engine \cite{unrealengine} or Unity \cite{unitygameengine}.
The generated scenarios can be rendered as photo-realistic images from arbitrary perspectives and with arbitrary scene content.
In the context of urban scenes one can design a street and populate it with cars, pedestrians, motorbikes, bicycles and different types of static objects.

Synthetic data has the advantage that almost any domain can be modelled, extended and adjusted to one's needs.
Further, even rare events can be modelled with required accuracy and variations that would be infeasible or too dangerous to acquire in real life.
An immense advantage of synthetic data is that annotations come for free.
Apart from a photo-realistic image, the 3D engine can also generate a depth map or a pixel-wise class annotation of the image since the type of the contained objects are known and defined by the modeller.

Certainly, there are domains where synthetic data generation is not possible or at least not as easy as in the case of urban scenes.
For example, synthetic images for medical applications might not depict a realistic distribution of certain types of diseases.
A network trained on such data could only perform well on artificial images and might not at all be able to detect pathological features in real images.

In this work we focus on the domains of urban and traffic scenes for synthetic data generation.
The goal is to review and compare convolutional neural network training scenarios that do not require large amounts of real-world data.
The question of whether or not training with synthetic images will improve the performance on real-world data is still valid.
It will be examined exemplary on several recent publications in that field.

\subsection{Datasets With Synthetic Images}

Several synthetic datasets have been composed in the recent years.
In this section we review some of the widely used datasets for urban scenes.
Experiments performed with these datasets and reported in literature will be reviewed in the next section.

\begin{figure}
 \centering
 \includegraphics[width=0.75\textwidth]{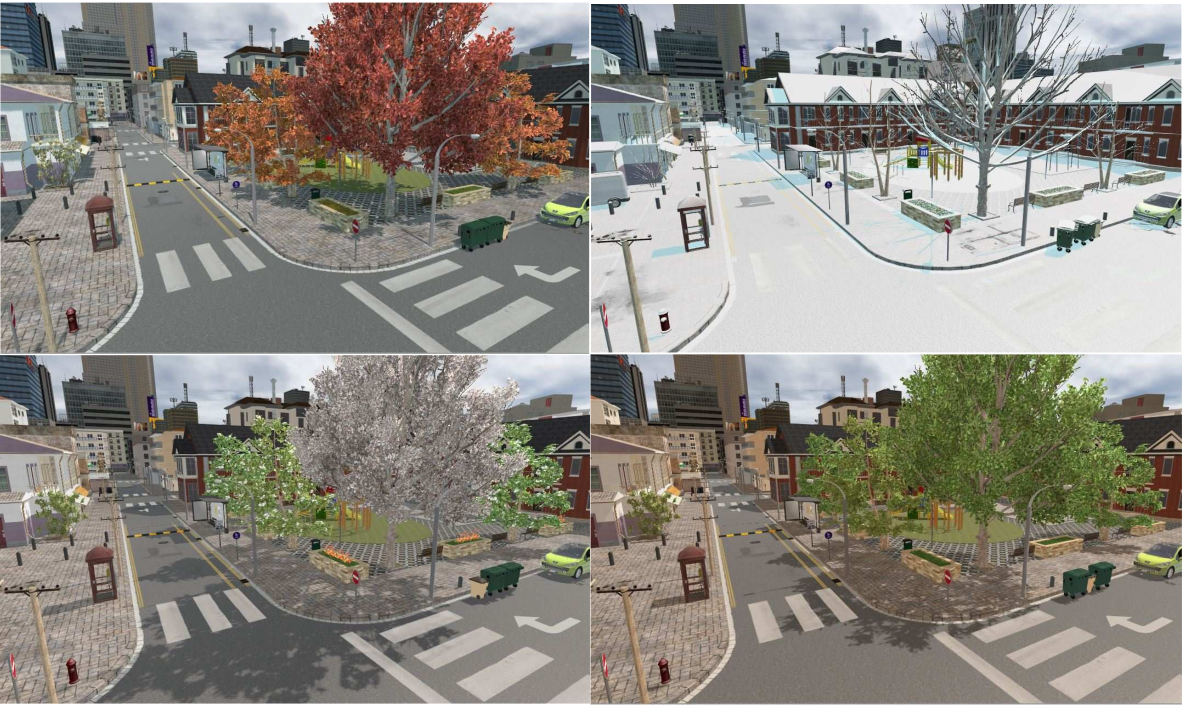}
 \caption{Example scene from SYNTHIA during the four different seasons. Image taken from \cite{ros2016synthia}.}
 \label{fig:synthia}
\end{figure}

Ros et al. present SYNTHIA (SYNTHetic collection of Imagery and Annotations of urban scenarios) \cite{ros2016synthia}, a photo-realistic dataset of a virtual city created with the Unity 3D engine.
It contains pixel-wise annotations for 13 object classes from multiple view points.
The images have a resolution of 960 $\times$ 720 pixels and horizontal field of view of 100 degrees.
SYNTHIA consists of two distinct image sets.
The first set provides around 13,000 images of random camera positions in the city.
The height of the virtual camera was fixed to be between 1.5 and 2 meters.
The second set provides four video sequences with around 50,000 images each of a simulated car ride across the city.
Each of the sequences depicts the same scenes, but during a different season (Figure\,\ref{fig:synthia}).

\begin{figure}
 \centering
 \includegraphics[width=0.75\textwidth]{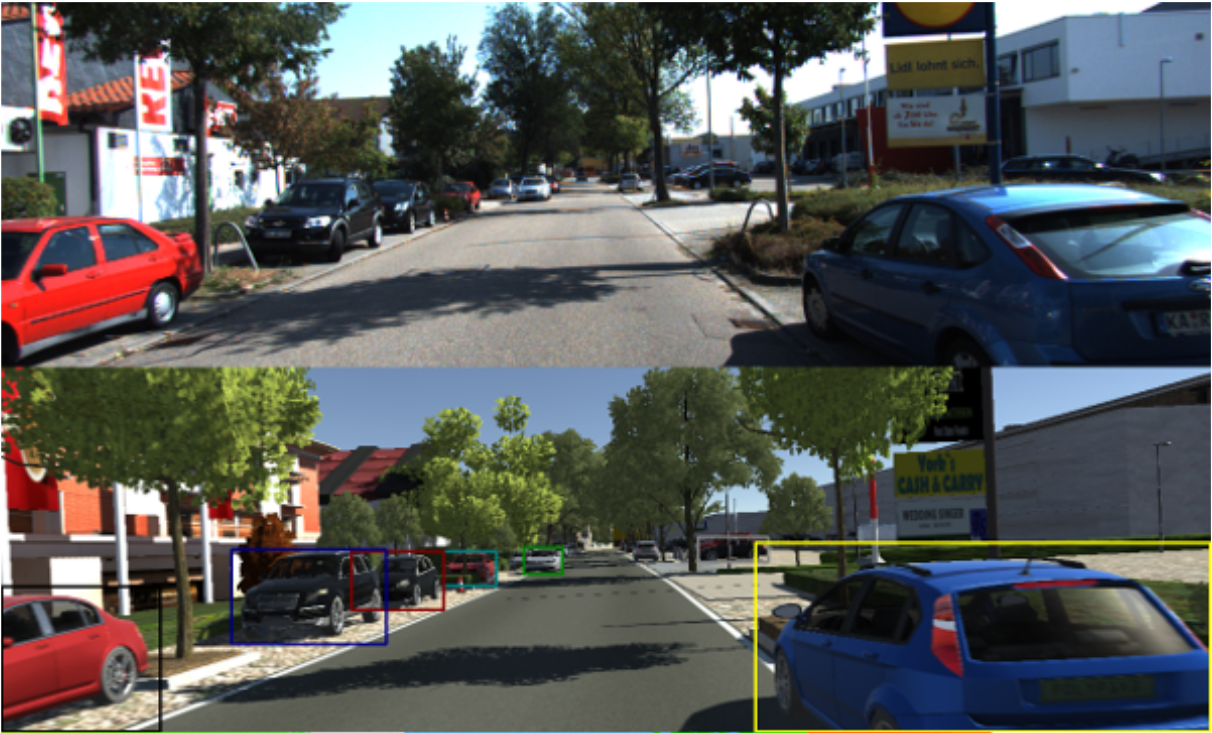}
 \caption{Example scene from KITTI (top) and the corresponding scene in VirtualKITTI (bottom). Image taken from \cite{gaidon2016virtual}.}
 \label{fig:virtualkitti}
\end{figure}

While SYNTHIA relies on a completely artificial city, Gaidon et al. present an approach to transfer an existing real-world dataset (KITTI \cite{geiger2013vision}) into the virtual domain, dubbed VirtualKITTI \cite{gaidon2016virtual} (Figure\,\ref{fig:virtualkitti}).
The resulting dataset is photo-realistic and densely labeled with an image resolution of 1242 $\times$ 375 pixels and a total of around 17,000 frames.
While the dataset is smaller than SYNTHIA, apart from object detection and segmentation, it provides annotations for other tasks, including tracking and optical flow.
In contrast to the real KITTI dataset, VirtualKITTI allows to modify the setup and conditions of the scenes.
Among others, this includes the number and position of objects and different lighting and weather conditions.

Johnson-Roberson et al., the creators of the Driving in the Matrix (DITM) dataset \cite{drivingmatrix} take a different approach.
Instead of creating a virtual environment they use a set of tools to capture photo-realistic images from a video game (GTA5, Rockstar games).
They focus on different vehicle classes for the tasks of object detection and segmentation.
The datasets consists of three distinct image sets with around 10,000, 50,000 and 200,000 images under different lighting and weather conditions.

The same video game is also used by Richter et al., the creators of the VIsual PERception benchmark (VIPER) \cite{richter2017playing}.
This dataset comprises a total of 254,000 full HD images with annotations for object detection, instance segmentation and optical flow.
Similar as DITM, this dataset provides images in different lighting and weather conditions, but does not solely focus on vehicles.
In a previous work \cite{richter2016playing} the authors released a predecessor of the VIPER dataset with only around 25,000 images that we will refer to as the GTA5 dataset.

Finally, the Virtual Environment for Instance Segmentation (VEIS) dataset described by Sadat Saleh et al. in \cite{sadat2018effective} focuses on instance segmentation of foreground objects.
The authors describe their dataset as ``not highly realistic'' regarding the textures of the objects, but with realistic shapes.
The dataset of about 61,000 images is split into a set of multi-class images with complex scenes and a single-class (multiple instance) images with simple scenes.

Some properties of the mentioned datasets are summarized in Table\,\ref{tab:overview_datasets}.
The authors of these datasets employ different training procedures which we will review below along with other approaches on synthetic datasets.

\begin{table}
\caption{Summary of some properties of the synthetic datasets mentioned in this work.}
\label{tab:overview_datasets}
\begin{center}
\begin{tabular}{lllc}
Dataset                  & \# images & Variations & Framework/Source \\
\hline
\hline
SYNTHIA \cite{ros2016synthia} & 13k, 50k       & seasons           & Unity\\
VirtualKITTI \cite{gaidon2016virtual} & 17k    & weather, lighting & Unity\\
DITM \cite{drivingmatrix}     & 10k, 50k, 200k & weather, lighting & GTA5\\
GTA5 \cite{richter2016playing}& 25k            & weather, lighting & GTA5\\
VIPER\cite{richter2017playing}& 254k           & weather, lighting & GTA5\\
VEIS \cite{sadat2018effective}& 61k            & multi/single class scenes & Unity
\end{tabular}
\end{center}
\end{table}

\subsection{Approaches Exploiting Synthetic Data}

The authors of the SYNTHIA dataset target the task of semantic segmentation in urban scenes.
They test the hypothesis that additional synthetic data aid in training neural networks with datasets containing real data.
In total they investigate four different datasets and two neural networks (T-NET \cite{ros2016training} and FCN \cite{long2015fully}).
Each of the networks is trained with each of the datasets with real images only.
In a second pass, the training is repeated with the same datasets that are mixed with synthetic images.
The training occurs with mixed batches of 10 images, four of which are synthetic.
For most classes the results of training with real and synthetic images are better than with real images only.
The authors observe an improvement for the network T-NET in average per-class and global accuracy.
On the other hand, the improvements for FCN only occur in average per-class accuracy.
Among the classes that gain most from additional synthetic images are classes that represent individual objects.
These are pedestrians, cyclists, but also cars and even poles.
Only small improvements or even worse results are observed for classes that occupy greater areas, such as sky, buildings, roads, vegetation.
The latter can be regarded as background classes, whereas the individual objects constitute the foreground of a scene.
This distinction will be discussed further below in this section.

The authors of VirtualKITTI primarily benchmark a tracking algorithm, however, following the tracking-by-detection paradigm.
Although their dataset provides more classes, in their publication they focus on cars only.
For their experiments they apply a Faster-R-CNN model \cite{ren2015faster} and replace the backbone by VGG16.
The network was pretrained on ImageNet.
The authors fine-tune it to cars from the Pascal VOC 2007 dataset.
In the last step transfer learning is applied with the KITTI and with the VirtualKITTI datasets, respectively.
Out of these two variants, the one trained on real data performs much better.
In a third experiment, transfer learning is again applied with the VirtualKITTI dataset, however, the resulting network is subsequently fine-tuned with the real data.
This last experiment improves over the training with real data, although by a smaller margin than the real data improves over synthetic data alone.

A different approach is reported by the authors of the DITM dataset.
They also focus on cars only and use Faster-R-CNN with VGG16.
In contrast to other approaches they completely separate real and synthetic data in training to better assess the results.
Training is performed on Cityscapes \cite{cordts2016cityscapes} and on different variants of their simulated data (with 10,000, 50,000 and 200,000 images).
Each of the 4 trained variants is subsequently evaluated on real data from the training set of the KITTI dataset.
The results were split into easy, moderate and hard to detect cars according to the KITTI dataset (the difficulty depends on the size of the car in the image, occlusion etc.).
In all three categories, training on Cityscapes leads to better results than training with 10,000 synthetic images, although Cityscapes provides only around 3,000 images for training.
However, increasing the synthetic image dataset to 50,000 beats training with Cityscapes by a large margin.
For easy to detect cars 50,000 images are sufficient and lead to best results.
For moderately and hard to detects cars the dataset with 200,000 leads to even better results.
The authors draw three conclusions from these experiments.
First, datasets with real images are too small to transfer the learned weights to other datasets.
Second, the current performance of neural network is limited by the available annotated training data rather than network architectures.
Finally, when training with synthetic images, the number of required training samples is much higher than with real images.
This indicates that real images have a higher variation in features, lighting, color etc. than simulated images.

The authors of the VIPER dataset \cite{richter2017playing} discussed above focus on a benchmarking suite for training on synthetic data.
In their paper they describe and measure statistical distributions of their synthetic dataset.
These include among others categories and instances per image, instances per semantic class and the distribution of distances of vehicles to the camera.
They find that the statistics of their proposed dataset closely resemble the statistics of Cityscapes.
In a subsequent evaluation they find that the relative performance of two investigated neural networks on their dataset is consistent with the performance of these networks on Cityscapes.
Unfortunately, in this work they make no comparison of training with real and synthetic images.
However, in a previous publication \cite{richter2016playing} they report such experiments.
Note that the synthetic dataset in the previous paper had around 25,000 images (GTA5 dataset), which is only around 10\% of the VIPER dataset.
In the previous work the authors conduct two distinct semantic segmentation experiments.
In the first experiment the authors explore gradually mixing real and synthetic images on the CamVid dataset \cite{brostow2009semantic}.
Training on real images leads to significantly better results than only using synthetic images.
However, training with all synthetic images, but only 25\% of all available real images leads to a performance close to training on all real images.
Increasing the amount of real images to 33\% already beats the results on real images.
Finally, taking all available real and synthetic training data clearly improves the results.
In the second experiment the authors train on the KITTI dataset and again observe an improvement when synthetic data is added in training.
In both experiments training is performed with mixed batches of four real and four synthetic images.

A common concern when training a network on synthetic images is the degrading performance on real images.
This performance gap is referred to as domain shift.
Sadat Saleh et al., the authors of the paper presenting the VEIS dataset \cite{sadat2018effective} do not explicitly address domain shift, but take it into account when designing their experiments.
They claim that foreground and background classes are affected differently by the domain shift and propose to treat them accordingly.
Sadat Saleh et al. argue that background classes such as road, sky, buildings and vegetation have realistic textures even in synthetic data.
On the other hand, only the shape, but not the texture of foreground classes such as cars, pedestrians and bicycles looks realistic.
The authors therefore propose to handle foreground classes in a detection-based manner to account for realistic shapes, while applying semantic segmentation to background classes because of their realistic texture.
We have already observed (an implicit) distinction between foreground and background classes in the results reported by the authors of the SYNTHIA dataset.
Mostly, adding synthetic data only improved the results for foreground classes.
Following the observations of Sadat Saleh et al. we conclude that the SYNTHIA results on foreground classes improved due to additional shape variations introduced with the synthetic data.
Sadat Saleh et al. train a VGG16-based DeepLab model \cite{chen2017deeplab} for semantic segmentation of background classes on the GTA5 dataset.
This dataset was chosen because of its photo-realistic textures of background classes.
Foreground classes are trained on the proposed VEIS dataset using the Mask-RCNN network and the results of both networks are fused.
The trained network is evaluated on the validation set of Cityscapes.
The proposed method is compared with training a single segmentation network on different synthetic datasets and consistently obtains better results.
The results are improved even further when the proposed approach is used to generate pseudo ground truth labels for unlabeled real images.

\subsection{Exemplary Approaches Explicitly Addressing Domain Shift} \label{examp-domain}

\begin{figure}
 \centering
 \includegraphics[width=0.9\textwidth]{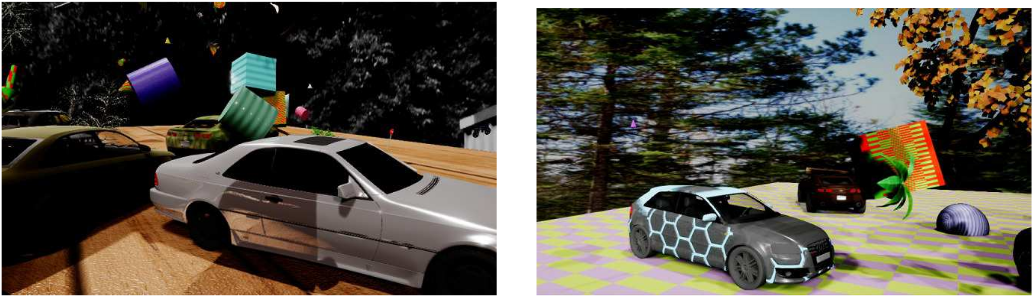}
 \caption{Examples for domain randomization. Image taken from \cite{tremblay2018training}.}
 \label{fig:domain_random}
\end{figure}

Several papers (for example \cite{drivingmatrix}, \cite{sadat2018effective}) report that synthetic data does not possess as much variation in appearance as real data.
Tremblay et al. \cite{tremblay2018training} address this shortcoming by an approach that belongs to the field of domain randomization.
The goal is to create synthetic images that do not look photo-realistic at all.
The authors generate images of cars with random, unrealistic parameters for lighting, pose and textures (Figure\,\ref{fig:domain_random}).
A total of 100,000 images are generated and used in the experiments.
The idea behind domain randomization is that the network has to learn to detect objects independently from their texture.
When presented with real images at test time the real texture is regarded as just another variation by the network.
An additional benefit is that time consuming creation of photo-realistic images is not necessary with domain randomization.
Tremblay et al. apply transfer learning with this randomized data on a pretrained network and use fine-tuning with real data.
They observe that it is beneficial to additionally update the weights of the earlier layers when applying transfer learning with synthetic data.
The reported results are better than with real images alone.

The authors of \cite{sankaranarayanan2018learning} propose to use GANs to decrease the domain shift.
In their approach synthetic data is not just used as additional training data.
Instead, a GAN is trained to distinguish between real and fake images - both, for the source domain of synthetic images and the target domain of real images.
An additional pair of networks is closely linked to the GAN during training and is used to train an image embedding that more closely resembles the target domain.
The presented experiments use SYNTHIA and GTA5 as source domains, respectively.
The target domain is Cityscapes.
The proposed approach leads to a significantly lower domain gap compared to a conventional training on synthetic images alone.
Still, training on real images is reported to lead to best results.
At first, this seems to contradict the other papers discussed above.
However, SYNTHIA and GTA5 are comparatively small datasets and in their respective papers were always mixed with real data to obtain an improvement.
We therefore assume that the reported results can be improved further with a large scale synthetic dataset.

\section{Outlook on Synthetic Data with GANs} \label{outlook}

Nowadays, a lot of innovation in the field of deep learning is happening the area of GANs.
Many different approaches for image to image translation have been proposed in the recent years.
A recurrent topic of image to image translation methods is translating semantic maps of urban scenes to photo-realistic images \cite{isola2017image}, \cite{chen2017photographic}, \cite{wang2018high}.
Such semantic maps highly resemble the ground truth annotation data of the Cityscapes dataset.
The work of Park et al. \cite{park2019semantic} not only translates semantic maps to photographs, but additionally allows to choose different styles for the resulting image.
This property covers both, style transfer (Section\,\ref{data-augment}) and domain randomization (Section\,\ref{examp-domain}).

Of course to massively apply such transformations from maps to photos one needs to create semantic maps. 
While such maps come from ground truth annotations in these papers, there have also been attempts to create complex semantic maps with GANs \cite{ghelfi2019adversarial} and to augment existing maps with inserted object instances \cite{lee2018context}.

With these recent research papers, we have all building blocks for new types of synthetic datasets.
One could synthesize semantic maps, augment the results by inserting required object instances and finally translate the resulting maps into photo-realistic images.
We expect so see such datasets in the near future and are excited about the possibilities these datasets will offer in the areas of training and augmenting with synthetic data.

\section{Summary} \label{summary}
In this work we reviewed several current approaches that aim at improving the performance of neural networks.
The presented ideas were twofold.
On one hand, we examined how small datasets can be used effectively to train powerful networks with limited amount of data.
On the other hand, we presented some recently published and successful experiments to train with real and synthetic data.
These experiments achieve better results than with real data alone.

We identify several guidelines for the development of computer vision applications based on neural networks.
First and most importantly, one should always consider using a pre-trained network, rather than training from scratch.
Second, in general it is recommended to apply transfer learning and fine-tuning.
This is especially the case if the total amount of training data is small.
Data augmentation techniques of various types have been proposed in literature and are a great tool to help a network to generalize better.

Synthetic data is easier to acquire and has a great potential to further improve the performance of neural networks.
However, the synthetic datasets that are available nowadays contain less variations than real data.
As a consequence, significantly more synthetic data is needed to achieve the same training results as with real data.
Nevertheless, mixing synthetic and real data has been shown to increase the performance of neural networks in several experiments.
Mixing is done either by training with mixed real and synthetic batches.
Another popular method is using synthetic data for transfer learning and later real data for fine-tuning.

The more one relies on synthetic data during training, the more domain shift becomes an issue.
Luckily, there is an indication that more synthetic data diminishes its effects and the resulting networks also perform well on real data.

For the task of semantic segmentation we saw that synthetic data might be enough to train background classes that comprise large image areas.
Contrary, synthetic data for foreground classes (i.e. classes that represent individual objects) has realistic shapes, but poor texture.
Thus, foreground classes benefit from a training procedure that aims at object detection, rather than semantic segmentation.
We would like to see more research on that topic to gain further insights.

Finally, GANs are a hot research topic and we expect more exciting datasets in the near future.
These could include automatically generated, photo-realistic urban scenes where relevant intances such as pedestrians or cars are automatically inserted in different poses at relevant positions in the images.

\bibliographystyle{splncs03.bst}
\bibliography{references}

\end{document}